%% file: neurips_2026.tex
\documentclass{article}

    \PassOptionsToPackage{numbers, compress}{natbib}

\usepackage[preprint]{neurips_2026}

\usepackage[utf8]{inputenc} 
\usepackage[T1]{fontenc}    
\usepackage{hyperref}       
\usepackage{url}            
\usepackage{booktabs}       
\usepackage{amsfonts}       
\usepackage{nicefrac}       
\usepackage{microtype}      
\usepackage{xcolor}         
\usepackage{amsmath}
\usepackage{algorithm}
\usepackage{algorithmic}

\usepackage{booktabs}
\usepackage{multirow}
\usepackage[table]{xcolor}
\usepackage{graphicx}

\title{NeuroOnline: Bridging Pretraining and Online Adaptation for EEG Foundation Models}

\author{%
Weibin Li$^{1}$\thanks{These authors contributed equally} \quad
Wendu Li$^{1}$\footnotemark[1] \quad
Yushan You$^{1}$ \quad
Chen Wei$^{1}$ \thanks{Corresponding authors} \quad
Quanying Liu$^{1}$\footnotemark[2] \\[6pt]
$^{1}$Southern University of Science and Technology \\
\texttt{liwb2023@mail.sustech.edu.cn, 12432838@mail.sustech.edu.cn, liuqy@sustech.edu.cn}
}

\begin{document}


\maketitle


\begin{abstract}
  EEG foundation models have shown strong potential in learning generalized representations across subjects and tasks. However, most existing approaches follow a pretraining–static deployment paradigm, which suffers from two key limitations: (1) misalignment between pretraining objectives and downstream tasks, and (2) limited adaptability to distribution shifts in online settings. We propose Online Neural Adaptation (NeuroOnline), a unified framework that enables continuous adaptation in online scenarios. NeuroOnline integrates two complementary mechanisms: (1) multi-view consistency learning, which enforces cross-view alignment to promote consistent and task-relevant representations, and (2) context-aware representation modulation, which leverages a learnable context prompt with cross-attention to dynamically adapt representations to evolving data distributions. Together, these mechanisms unify representation alignment and dynamic adaptation. Experiments on multiple EEG benchmarks show that NeuroOnline consistently outperforms strong baselines in online settings, achieving better performance under distribution shifts. Ablation and sensitivity studies further validate the necessity of each component and the effectiveness of the overall design.
\end{abstract}

\section{Introduction}

Electroencephalography (EEG) is a key modality for brain-computer interfaces (BCIs), enabling characterization of neural activity for applications such as communication, control, cognitive monitoring, and neuromodulation \cite{abiri2019comprehensive, aggarwal2022review, lotte2018review, varbu2022past}. Inspired by the success of large-scale pretraining in vision and natural language processing \cite{10.1145/3605943, 9879206, chen2020simple, devlin2019bert}, recent studies have explored EEG foundation models to learn generalizable representations across subjects, tasks, and recording conditions \cite{kuruppu2026eeg, lai2025simple}. These models exhibit strong cross-scenario generalization. This makes them promising candidates for general-purpose backbones in a wide range of downstream brain signal decoding tasks \cite{wu2025adabrain, https://doi.org/10.1155/2020/8875426, BOONYAKITANONT2020101702, SHARMA2022116634, AMIN2019542, 10.1007/978-3-030-21642-9_8, 7478117, e18090272}.

\begin{figure}[t]
\begin{center}
\centerline{\includegraphics[width=\columnwidth]{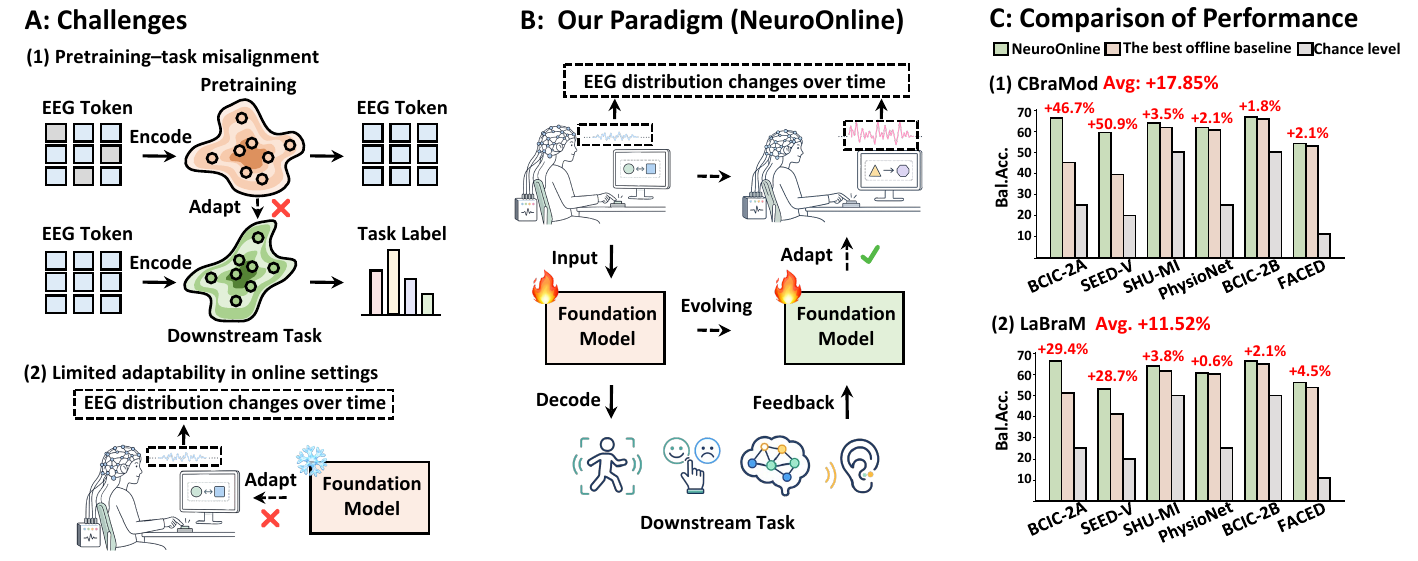}}
\caption{\textbf{Motivation.} (A) Challenges: pretraining–task misalignment and limited adaptability under distribution shifts in online EEG settings. (B) Our paradigm (NeuroOnline): continual adaptation of foundation models to evolving neural dynamics. 
(C) Performance comparison: NeuroOnline consistently achieves relative improvements over the strongest offline baseline (off-FT-MLP) across multiple datasets and EEG foundation models in online settings.}
\label{fig1}
\end{center}
\end{figure}

Despite recent progress, most EEG foundation models still adopt a pretraining–static deployment paradigm, which faces two fundamental challenges (Fig. \ref{fig1}A): \textbf{(1) Pretraining–task misalignment}. Self-supervised objectives (e.g., masked reconstruction) inherently differ from downstream decoding goals, resulting in a mismatch between learned representations and task-specific data distributions \cite{9462394, wang2024cbramod, ICLR2024_47393e85}. \textbf{(2) Limited adaptability under dynamic distribution shifts in online settings}. EEG signals are inherently non-stationary, with statistical properties that evolve over time, leading to distribution shifts during online deployment \cite{cecotti2025non, shen2019challenge, RAZA2019154}. In such settings, models must update their representations as new data arrive. However, without adaptive mechanisms, static models cannot adjust accordingly, further exacerbating the misalignment.



To address the above challenges, we propose a new learning paradigm, Online Neural Adaptation (NeuroOnline), that enables foundation models to continuously adapt to neural dynamics in online settings (Fig. \ref{fig1}B). NeuroOnline consists of two key mechanisms: \textbf{(1) Multi-view consistency learning}. We leverage observed online samples to generate multiple augmented views through stochastic transformations, such as temporal and frequency masking, and enforce consistency across them. This mechanism enables the model to learn consistent and task-relevant representations, thereby mitigating the distribution mismatch between pretraining and downstream tasks. \textbf{(2) Context-aware representation modulation.} We utilize a learnable context prompt to capture latent information from online data and employ cross-attention to dynamically adjust representations, enabling continuous adaptation to evolving data distributions. These mechanisms correspond to representation alignment and dynamic adaptation, respectively, thereby bridging the gap between pretraining and online settings (Fig. \ref{fig1}). Our main contributions are summarized as follows:




\begin{itemize}
    \item \textbf{Paradigm Innovation.} We propose Online Neural Adaptation (NeuroOnline), which shifts EEG foundation models from a static setting to continual adaptation, bridging the gap between pretraining and online settings.
    \item \textbf{Methodological Innovation.} We propose a multi-view consistency learning mechanism for pretraining–task alignment and a context-aware modulation mechanism for continuous adaptation under evolving data distributions.
    \item \textbf{Extensive Experiments.} We conduct comprehensive experiments across multiple foundation models and diverse EEG datasets, showing that NeuroOnline consistently outperforms strong baselines in online settings, with ablation and sensitivity analyses validating the effectiveness of its components.
\end{itemize}

\section{Related Work}

\subsection{EEG Foundation Models}

Existing EEG foundation models aim to learn generalizable neural representations across subjects, tasks, and recording conditions \cite{kuruppu2026eeg, lai2025simple, wu2025adabrain}. This improves generalization in downstream brain signal decoding. Most approaches follow a paradigm of large-scale self-supervised pretraining with Transformer-based architectures to capture spatio-temporal dependencies. Representative methods include CBraMod \cite{wang2024cbramod}, LaBraM \cite{ICLR2024_47393e85}, EEGPT \cite{NEURIPS2024_4540d267}, and BIOT \cite{NEURIPS2023_f6b30f3e}, which differ in model design and training strategies but share the goal of learning scalable and transferable EEG representations. For example, CBraMod \cite{wang2024cbramod} adopts a criss-cross Transformer with parallel spatio-temporal attention and masked pretraining to learn scalable EEG representations, while LaBraM \cite{ICLR2024_47393e85} leverages vector-quantized tokenization and masked pretraining on large-scale unlabeled data to learn transferable representations. \textbf{However, these models are largely built upon a pretraining–static deployment paradigm. Their parameters remain fixed after deployment, with no mechanisms for continuous adaptation in online settings. As neural data distributions evolve over time, these models struggle to adapt effectively.}



\subsection{Online Learning}
Online learning has proven effective in computer vision and natural language processing, where continuously updating models with incoming data mitigates distribution shift and improves robustness in dynamic environments \cite{HOI2021249, 9880363, huang2025online, 10.1145/3735633, wang2020tent}. Inspired by this success, extending online learning to EEG representation learning provides a promising direction. \textbf{However, EEG signals exhibit stronger non-stationarity, greater inter-subject variability, and more complex temporal dynamics than image or text data \cite{9508768, du2022eeg}, leading to more severe and harder-to-model distribution shifts in online settings. Consequently, developing dedicated online adaptation paradigms for EEG foundation models remains an open challenge.} To address this, we propose Online Neural Adaptation (NeuroOnline), which integrates the multi-view consistency learning mechanism with the context-aware representation modulation mechanism to enable continuous adaptation in online environments.

\section{Method}

\subsection{Problem Formulation and Overview}

Let $\{X_t\}_{t=1}^T$ denote a stream of sequentially arriving EEG segments, where each $X_t \in \mathbb{R}^{m \times N}$ represents a short temporal window with $m$ frames and $N$ channels. At step $t$, the model receives the current observation $X_t$ and predicts the corresponding task label $y_t \in \mathbb{R}^d$, where $d$ is the number of classes. Existing EEG foundation models typically follow a pretraining–static deployment paradigm, which suffers from two key limitations: (1) pretraining–task misalignment and (2) limited adaptability under distribution shifts in online settings. Our goal is to predict $y_t$ accurately while continuously adapting the model to evolving data distributions. 

\begin{figure}[t]
\begin{center}
\centerline{\includegraphics[width=\columnwidth]{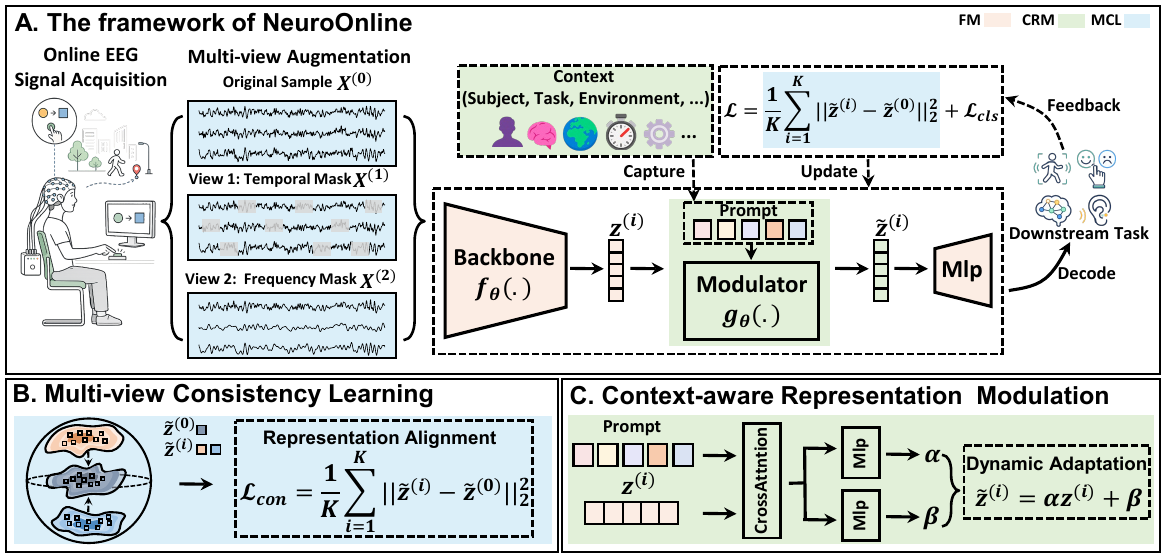}}
\caption{\textbf{Framework of NeuroOnline.} (A) Overall pipeline: given streaming EEG data, NeuroOnline constructs multi-view augmentations, encodes representations with a backbone model, and integrates multi-view consistency learning (MCL) and context-aware representation modulation (CRM) for continuous online adaptation. (B) Multi-view consistency learning: representations from augmented views are aligned to mitigate pretraining–task misalignment. (C) Context-aware representation modulation: representations are dynamically modulated via contextual prompts to adapt to evolving data distributions.}
\label{fig2}
\end{center}
\end{figure}

We present NeuroOnline, a unified framework for online adaptation of EEG foundation models, as illustrated in Fig.~\ref{fig2}. The framework comprises two key components: (i) multi-view consistency learning, which enforces cross-view alignment to learn consistent and task-relevant representations, and (ii) context-aware representation modulation, which modulates representations via context-conditioned scaling and shifting driven by prompt–representation interactions. Together, these components enable a unified online learning framework that mitigates pretraining–task misalignment and distribution shifts, thereby bridging the gap between pretraining and online adaptation.

\subsection{Multi-view Consistency Learning (MCL)}
To mitigate the misalignment between pretraining objectives and downstream tasks, we propose the multi-view consistency learning mechanism that enforces representation alignment across augmented views, enabling consistent and task-relevant feature learning.

\textbf{Multi-view Construction.} We construct multiple augmented views from online samples observed up to time step $t$, ensuring a strictly causal setting. For each sample $X$, we define it as the anchor view $X^{(0)}$ and generate $K$ augmented views $\{X^{(i)}\}_{i=1}^K$ via stochastic transformations, including temporal masking and frequency masking, where $K=2$. Specifically, temporal masking randomly removes contiguous segments along the time dimension to simulate local temporal corruption, while frequency masking suppresses selected frequency bands to perturb spectral structure. These augmentations preserve semantic content while introducing complementary perturbations, enabling the model to learn consistent and task-relevant representations.

\textbf{Consistency Learning.}
Given the constructed views, we enforce consistency in the modulated representation space. For each sample $X$, we compute the modulated representations for all views as:
\begin{equation}
\label{modulated representations}
\tilde{z}^{(i)} = g_\theta\big(f_\theta(X^{(i)})\big), \quad i = 0, \dots, K,
\end{equation}
where $f_\theta$ denotes the backbone model and $g_\theta$ the context-aware modulator, with $i=0$ corresponding to the anchor view. We enforce consistency by aligning each augmented view with the anchor representation $\tilde{z}^{(0)}$ (Fig. \ref{fig2}B), with the loss defined as:
\begin{equation}
\label{lcons}
\mathcal{L}_{\text{cons}} = \frac{1}{K} \sum_{i=1}^{K} \left\| \tilde{z}^{(i)} - \tilde{z}^{(0)} \right\|_2^2.
\end{equation}
By imposing consistency in the modulated representation space, the model adapts alignment to evolving data distributions. This promotes view-consistent and task-relevant representations, mitigating pretraining–task mismatch.

\subsection{Context-aware Representation Modulation (CRM)}
To adapt to evolving data distributions in online environments, we propose a context-aware representation modulation mechanism that adapts representations by leveraging latent contextual information.

\textbf{Context Prompt.}
We maintain a learnable prompt $\mathcal{P}$ to capture latent contextual information in the online data stream. The prompt is dynamically activated through interaction with the input, allowing it to adapt to evolving data distributions. 

\textbf{Context-aware Modulation.}
Given the input representation $z = f_\theta(X)$, we adaptively modulate it via a context-aware function $g_\theta$. Specifically, $g_\theta$ consists of a cross-attention module followed by two parallel MLP heads. The attention mechanism captures interactions between the input representation and the prompt $\mathcal{P}$, while the two MLP heads output the scaling and shifting parameters $\alpha$ and $\beta$, respectively. The modulated representation is then:
\begin{equation}
\label{modulation}
\tilde{z} = g_\theta(z, \mathcal{P}) = \alpha \odot z + \beta,
\end{equation}
where $\odot$ denotes element-wise multiplication. The same modulation is applied consistently to all augmented views (see Fig. \ref{fig2}C for details). By incorporating the contextual information into the representation space, this mechanism enables adaptive modulation under distribution shifts, leading to improved discriminability and robustness.

\subsection{Training and Optimization}

We train the model in an online setting. The model is updated incrementally based on observed samples, ensuring a causal setting. The overall objective consists of a classification loss and a consistency loss:
\begin{equation}
\label{total-loss}
    \mathcal{L} = \mathcal{L}_{\text{cls}} + \lambda \mathcal{L}_{\text{cons}},
\end{equation}
where $\mathcal{L}_{\text{cls}}$ denotes the task-specific classification loss (e.g., binary cross-entropy with logits for binary classification, or cross-entropy loss for multi-class classification), $\mathcal{L}_{\text{cons}}$ enforces consistency across multiple views, and $\lambda$ balances the two objectives. As new samples arrive, the model is continuously updated, enabling adaptation to evolving data distributions. The overall optimization procedure is summarized in Algorithm~\ref{algorithm}.

\section{Experiments}
\subsection{Experimental setup}


\textbf{Datasets:} We evaluate the proposed method on six widely used EEG decoding benchmarks across two downstream tasks: (1) Motor Imagery Classification, including BCIC-2A \cite{brunner2008bci}, BCIC-2B \cite{steyrl2016random}, SHU-MI \cite{ma2022large}, and PhysioNet-MI \cite{goldberger2000physiobank, schalk2004bci2000}; and (2) Emotion Recognition, including FACED \cite{chen2023large} and SEED-V \cite{liu2021comparing}. A complete list of tasks and corresponding datasets is provided in Table \ref{tab:bci_datasets} and Appendix \ref{Additional Details on Datasets and Preprocessing}. To better reflect real-world online settings, preprocessing is performed on individual EEG segments (i.e., epochs or trials) rather than on entire continuous EEG recordings (see Appendix \ref{Additional Details on Datasets and Preprocessing} for details). For each dataset, following the CBraMod \cite{wang2024cbramod}, we split the data into training, validation, and test sets based on its original protocol or data structure (see Appendix \ref{Additional Details on Datasets and Preprocessing} for details). The training set is used for offline training or fine-tuning of baseline models, the validation set for hyperparameter selection, and the test set is streamed sequentially to simulate a realistic online setting, where we evaluate the model’s continual adaptation capability. 

\begin{table}[h]
\centering
\caption{Overview of downstream BCI tasks and datasets.}
\label{tab:bci_datasets}
\resizebox{\linewidth}{!}{
\begin{tabular}{l l c c c c c}
\toprule
\textbf{BCI Tasks} & \textbf{Datasets} & \textbf{Channels} & \textbf{Duration} & \textbf{Samples} & \textbf{Label} \\

\midrule
I. Motor Imagery Classification
& BCIC-2A & 22 & 4s & 5,088 & 4-class \\
& BCIC-2B & 3 & 4s & 6,520 & 2-class \\
& SHU-MI & 32 & 4s & 11,988 & 2-class \\
& PhysioNet-MI & 64 & 4s & 9,837 & 4-class \\

\midrule
II. Emotion Recognition 
& FACED & 32 & 10s & 10,332 & 9-class \\
& SEED-V & 62 & 1s & 115,001 & 5-class \\

\bottomrule
\end{tabular}
}
\end{table}

\textbf{Baselines:} Existing studies on online continual adaptation for EEG foundation models remain limited, and standardized baselines for this setting are largely absent. To enable a comprehensive evaluation, we construct two representative groups of baselines. The first group consists of \textbf{offline methods}, where models are fine-tuned on the training set and directly applied to the test set without any online updates, including: (i) LoRA-based fine-tuning with a linear head (off-LoRA-LIN) \cite{hu2022lora}, (ii) LoRA-based fine-tuning with an MLP head (off-LoRA-MLP), (iii) full fine-tuning with a linear head (off-FT-LIN), and (iv) full fine-tuning with an MLP head (off-FT-MLP). The second group consists of \textbf{online updating methods}. Although dedicated online learning approaches for EEG foundation models are not yet established, we construct a practical online baseline by adapting model parameters during online evaluation. Specifically, we adopt a simplified variant of our proposed method (denoted as NeuroOnline.a), where the multi-view consistency learning (MCL) and context-aware representation modulation (CRM) modules are removed, retaining only the basic parameter updating mechanism. By comparing offline and online baselines, we systematically evaluate the adaptation capability of our method and its relative advantages in online environments (More details are provided in Appendix \ref{Additional Details on Baselines}).


\textbf{Backbone Models:} We evaluate the proposed method on four representative EEG foundation models, including CBraMod \cite{wang2024cbramod}, LaBraM \cite{ICLR2024_47393e85}, EEGPT \cite{NEURIPS2024_4540d267}, and BIOT \cite{NEURIPS2023_f6b30f3e} (More details are provided in Appendix \ref{More Details for Backbone Models}). Without modifying their original architectures, we incorporate the proposed online adaptation mechanism into these models to assess its generalization and effectiveness across different backbones.

\textbf{Metrics:} For binary classification, we report Balanced Accuracy (Bal.Acc.), AUC-PR, and AUROC, with AUROC used as the monitor metric. For multi-class classification, we report Balanced Accuracy (Bal.Acc.), Cohen’s Kappa (Kappa), and Weighted F1 (F1-w), with Cohen’s Kappa as the monitor metric (More details are provided in Appendix \ref{Details of Evaluation Metrics}). All results are averaged over five random seeds and reported with standard deviations. $\Delta IMP \uparrow$ denotes the percentage relative improvement over the best competing method for each metric. *** indicates statistical significance assessed using the Wilcoxon signed-rank test ($p < 0.001$).

\textbf{Implementation Details:} All experiments are implemented in Python 3.10.20 using PyTorch (version 2.11.0, CUDA 12.6). Experiments are conducted on a NVIDIA GeForce RTX 4090 GPU (24 GB memory). Detailed hyperparameter settings are provided in Appendix \ref{The Details of hyperparameter settings}.

\begin{table*}[t]
\centering
\caption{Performance comparison with baselines on motor imagery classification benchmarks under online settings.}
\label{tab:motor_imagery_results}
\resizebox{\textwidth}{!}{
\begin{tabular}{lcccccc}
\toprule
& \multicolumn{6}{c}{\textbf{CBraMod}} \\
\cmidrule(lr){1-7}
& \multicolumn{3}{c}{\textbf{BCIC-2A, 4-class}} 
& \multicolumn{3}{c}{\textbf{SHU-MI, 2-class}} \\
\cmidrule(lr){2-4} \cmidrule(lr){5-7}
\textbf{Methods}
& \textbf{Bal.Acc.}
& \textbf{Kappa} 
& \textbf{F1-w}
& \textbf{Bal.Acc.}
& \textbf{AUC-PR} 
& \textbf{AUROC} \\

\midrule
off-LoRA-LIN 
& 29.24 $\pm$ 2.31 
& 5.65 $\pm$ 3.09 
& 17.27 $\pm$ 3.40
& 62.86 $\pm$ 0.25 
& 69.21 $\pm$ 0.75 
& 69.03 $\pm$ 0.54 \\

off-LoRA-MLP 
& 30.54 $\pm$ 1.26 
& 7.38 $\pm$ 1.68 
& 19.21 $\pm$ 1.48 
& 63.01 $\pm$ 0.69 
& 68.55 $\pm$ 0.49 
& 68.54 $\pm$ 0.61 \\

off-FT-LIN 
& 44.55 $\pm$ 4.11 
& 26.06 $\pm$ 5.47 
& 39.88 $\pm$ 5.47 
& 61.55 $\pm$ 1.09 
& 66.93 $\pm$ 1.53 
& 66.67 $\pm$ 1.69 \\

off-FT-MLP 
& 45.12 $\pm$ 4.75 
& 26.83 $\pm$ 6.34 
& 40.46 $\pm$ 7.17 
& 61.69 $\pm$ 0.75 
& 68.55 $\pm$ 1.31 
& 68.06 $\pm$ 1.81 \\

\midrule

NeuroOnline.a 
& 62.62 $\pm$ 1.52 
& 50.16 $\pm$ 2.02 
& 62.43 $\pm$ 1.63 
& 63.39 $\pm$ 0.98 
& 68.53 $\pm$ 0.92 
& 68.53 $\pm$ 1.08 \\

\midrule

\rowcolor{blue!15}
\textbf{NeuroOnline***}
& \textbf{66.18} $\pm$ 2.11
& \textbf{54.91} $\pm$ 2.81
& \textbf{66.02} $\pm$ 2.11
& \textbf{63.86} $\pm$ 1.13
& \textbf{69.26} $\pm$ 0.81
& \textbf{69.41} $\pm$ 0.81 \\

\rowcolor{blue!15}
$\Delta IMP \uparrow$
& \textbf{+5.69\%} 
& \textbf{+9.47\%} 
& \textbf{+5.75\%} 
& \textbf{+0.74\%} 
& \textbf{+0.07\%} 
& \textbf{+0.55\%} \\

\toprule
& \multicolumn{3}{c}{\textbf{PhysioNet-MI, 4-class}}
& \multicolumn{3}{c}{\textbf{BCIC-2B, 2-class}} \\
\cmidrule(lr){2-4} \cmidrule(lr){5-7}
\textbf{Methods}
& \textbf{Bal.Acc.}
& \textbf{Kappa} 
& \textbf{F1-w}
& \textbf{Bal.Acc.}
& \textbf{AUC-PR} 
& \textbf{AUROC} \\

\midrule
off-LoRA-LIN 
& 58.20 $\pm$ 0.85 
& 44.26 $\pm$ 1.14 
& 57.99 $\pm$ 0.85 
& 50.35 $\pm$ 0.82 
& 69.08 $\pm$ 1.49 
& 69.56 $\pm$ 2.45  \\

off-LoRA-MLP 
& 58.03 $\pm$ 1.16 
& 44.04 $\pm$ 1.55 
& 58.05 $\pm$ 1.15 
& 60.01 $\pm$ 1.23 
& 64.38 $\pm$ 1.70 
& 65.18 $\pm$ 1.17  \\

off-FT-LIN 
& 60.58 $\pm$ 0.88 
& 47.44 $\pm$ 1.17 
& 60.36 $\pm$ 0.85 
& 66.14 $\pm$ 1.16 
& 69.42 $\pm$ 1.22 
& 71.87 $\pm$ 1.08  \\

off-FT-MLP 
& 60.46 $\pm$ 0.41 
& 47.28 $\pm$ 0.55 
& 60.36 $\pm$ 0.53 
& 65.59 $\pm$ 1.58 
& 68.99 $\pm$ 1.13 
& 71.44 $\pm$ 1.28  \\

\midrule
NeuroOnline.a 
& 53.89 $\pm$ 1.29 
& 38.52 $\pm$ 1.73 
& 53.89 $\pm$ 1.36 
& 64.86 $\pm$ 1.59 
& 71.66 $\pm$ 1.16 
& 71.62 $\pm$ 1.00  \\

\midrule
\rowcolor{blue!15}
\textbf{NeuroOnline***} 
& \textbf{61.74} $\pm$ 0.86
& \textbf{48.99} $\pm$ 1.15
& \textbf{61.76} $\pm$ 0.85
& \textbf{66.74} $\pm$ 0.61
& \textbf{72.63} $\pm$ 1.55
& \textbf{73.04} $\pm$ 1.29 \\

\rowcolor{blue!15}
$\Delta IMP \uparrow$
& \textbf{+1.91\%} 
& \textbf{+3.27\%} 
& \textbf{+2.32\%} 
& \textbf{+0.91\%} 
& \textbf{+1.35\%} 
& \textbf{+1.63\%} \\

\toprule
& \multicolumn{6}{c}{\textbf{LaBraM}} \\
\cmidrule(lr){1-7}
& \multicolumn{3}{c}{\textbf{BCIC-2A, 4-class}} 
& \multicolumn{3}{c}{\textbf{SHU-MI, 2-class}} \\
\cmidrule(lr){2-4} \cmidrule(lr){5-7}
\textbf{Methods}
& \textbf{Bal.Acc.}
& \textbf{Kappa} 
& \textbf{F1-w}
& \textbf{Bal.Acc.}
& \textbf{AUC-PR} 
& \textbf{AUROC} \\

\midrule
off-LoRA-LIN 
& 46.37 $\pm$ 2.81 
& 28.50 $\pm$ 3.74 
& 44.22 $\pm$ 3.26 
& 62.59 $\pm$ 0.21 
& 67.75 $\pm$ 0.27 
& 68.31 $\pm$ 0.49 \\

off-LoRA-MLP 
& 46.09 $\pm$ 1.31 
& 28.12 $\pm$ 1.75 
& 44.06 $\pm$ 1.74 
& 61.62 $\pm$ 1.37 
& 67.35 $\pm$ 1.19 
& 67.52 $\pm$ 1.57 \\

off-FT-LIN 
& 47.40 $\pm$ 2.17 
& 29.86 $\pm$ 2.89 
& 45.25 $\pm$ 2.88 
& 60.66 $\pm$ 1.30 
& 65.78 $\pm$ 1.49 
& 65.36 $\pm$ 1.54 \\

off-FT-MLP 
& 51.13 $\pm$ 1.47 
& 34.84 $\pm$ 1.96 
& 49.75 $\pm$ 1.83 
& 61.59 $\pm$ 0.99 
& 68.43 $\pm$ 1.98 
& 67.16 $\pm$ 2.10 \\

\midrule
NeuroOnline.a 
& 62.08 $\pm$ 2.33 
& 49.44 $\pm$ 3.11 
& 61.97 $\pm$ 2.36 
& 62.99 $\pm$ 0.66 
& 69.22 $\pm$ 0.72 
& 69.15 $\pm$ 0.96 \\

\midrule
\rowcolor{blue!15}
\textbf{NeuroOnline***} 
& \textbf{66.18} $\pm$ 0.85
& \textbf{54.91} $\pm$ 1.13 
& \textbf{66.03} $\pm$ 0.86
& \textbf{63.92} $\pm$ 0.52
& \textbf{70.17} $\pm$ 0.87
& \textbf{70.16} $\pm$ 0.50 \\

\rowcolor{blue!15}
$\Delta IMP \uparrow$
& \textbf{+6.60\%} 
& \textbf{+11.06\%} 
& \textbf{+6.55\%} 
& \textbf{+1.48\%} 
& \textbf{+1.37\%} 
& \textbf{+1.46\%} \\

\toprule
& \multicolumn{3}{c}{\textbf{PhysioNet-MI, 4-class}}
& \multicolumn{3}{c}{\textbf{BCIC-2B, 2-class}} \\
\cmidrule(lr){2-4} \cmidrule(lr){5-7}
\textbf{Methods}
& \textbf{Bal.Acc.}
& \textbf{Kappa} 
& \textbf{F1-w}
& \textbf{Bal.Acc.}
& \textbf{AUC-PR} 
& \textbf{AUROC} \\

\midrule
off-LoRA-LIN 
& 52.79 $\pm$ 1.08 
& 37.04 $\pm$ 1.44 
& 52.80 $\pm$ 1.14 
& 66.12 $\pm$ 1.37 
& 71.00 $\pm$ 2.82 
& 72.77 $\pm$ 1.87  \\

off-LoRA-MLP 
& 56.95 $\pm$ 0.65 
& 42.60 $\pm$ 0.86 
& 57.01 $\pm$ 0.71 
& 65.78 $\pm$ 2.16 
& 71.16 $\pm$ 3.07 
& 72.43 $\pm$ 2.21  \\

off-FT-LIN 
& 58.25 $\pm$ 1.18 
& 44.33 $\pm$ 1.58 
& 58.23 $\pm$ 1.22 
& 65.00 $\pm$ 0.80 
& 68.38 $\pm$ 0.68 
& 71.05 $\pm$ 0.61  \\

off-FT-MLP 
& 60.14 $\pm$ 0.60 
& 46.85 $\pm$ 0.81 
& 60.27 $\pm$ 0.53 
& 64.92 $\pm$ 1.59 
& 67.45 $\pm$ 2.56 
& 70.24 $\pm$ 2.52  \\

\midrule
NeuroOnline.a 
& 54.38 $\pm$ 0.67 
& 39.17 $\pm$ 0.89 
& 54.37 $\pm$ 0.67 
& 66.27 $\pm$ 1.59 
& 72.09 $\pm$ 1.43 
& 72.25 $\pm$ 1.96  \\

\midrule
\rowcolor{blue!15}
\textbf{NeuroOnline***} 
& \textbf{60.53} $\pm$ 0.80
& \textbf{47.38} $\pm$ 1.06
& \textbf{60.50} $\pm$ 0.94
& \textbf{66.28} $\pm$ 1.82
& \textbf{73.28} $\pm$ 1.04
& \textbf{73.11} $\pm$ 0.88 \\

\rowcolor{blue!15}
$\Delta IMP \uparrow$
& \textbf{+0.65\%} 
& \textbf{+1.13\%} 
& \textbf{+0.38\%} 
& \textbf{+0.02\%} 
& \textbf{+1.65\%} 
& \textbf{+0.47\%} \\

\bottomrule
\end{tabular}
}
\end{table*}

\subsection{Performance Comparison with Baselines}



\textbf{Motor Imagery:} As shown in Table \ref{tab:motor_imagery_results}, we evaluate the proposed method on BCIC-2A \cite{brunner2008bci}, BCIC-2B \cite{steyrl2016random}, SHU-MI \cite{ma2022large}, and PhysioNet-MI \cite{goldberger2000physiobank, schalk2004bci2000} using two representative EEG foundation models: CBraMod \cite{wang2024cbramod} and LaBraM \cite{ICLR2024_47393e85}. We compare against both offline and online baselines, with statistical significance assessed via the Wilcoxon signed-rank test. Across different datasets and evaluation metrics, NeuroOnline consistently achieves the best overall performance and significantly outperforms the strongest baseline ($p < 0.001$). Notably, the gains are more pronounced in multi-class settings, where Kappa shows an average relative improvement of $6.23\%$.


\textbf{Emotion Recognition:} As shown in Table \ref{tab:emotion_recognition_results}, we further evaluate the method on FACED \cite{chen2023large} and SEED-V \cite{liu2021comparing} under the same backbone settings. Our approach consistently achieves the best performance on most datasets and metrics, and yields statistically significant improvements over the strongest baselines ($p < 0.001$). Notably, in the more challenging emotion recognition scenario, our method maintains consistent gains across metrics, with an average relative improvement of 19.44\% (15.84\% for Bal.Acc., 27.54\% for Kappa, and 14.94\% for F1-w).

\begin{table*}[h]
\centering
\caption{Performance comparison with baselines on emotion recognition benchmarks under online settings.}
\label{tab:emotion_recognition_results}
\resizebox{\textwidth}{!}{
\begin{tabular}{lcccccc}
\toprule
& \multicolumn{6}{c}{\textbf{CBraMod}} \\
\cmidrule(lr){1-7}
& \multicolumn{3}{c}{\textbf{FACED, 9-class}} 
& \multicolumn{3}{c}{\textbf{SEED-V, 5-class}} \\
\cmidrule(lr){2-4} \cmidrule(lr){5-7}
\textbf{Methods}
& \textbf{Bal.Acc.}
& \textbf{Kappa} 
& \textbf{F1-w}
& \textbf{Bal.Acc.}
& \textbf{Kappa} 
& \textbf{F1-w} \\

\midrule
off-LoRA-LIN 
& 41.16 $\pm$ 0.89 
& 33.55 $\pm$ 1.00 
& 41.07 $\pm$ 0.95 
& 37.88 $\pm$ 0.20 
& 22.52 $\pm$ 0.24 
& 38.15 $\pm$ 0.25 \\

off-LoRA-MLP 
& 52.67 $\pm$ 0.47 
& 46.29 $\pm$ 0.53 
& 52.09 $\pm$ 0.38 
& 40.07 $\pm$ 0.60 
& 25.71 $\pm$ 0.86 
& 41.05 $\pm$ 0.69 \\

off-FT-LIN 
& 42.15 $\pm$ 0.75 
& 34.56 $\pm$ 0.91 
& 41.89 $\pm$ 0.96 
& 37.40 $\pm$ 0.25 
& 21.92 $\pm$ 0.27
& 37.98 $\pm$ 0.26 \\

off-FT-MLP 
& 52.90 $\pm$ 0.39 
& 46.67 $\pm$ 0.40 
& 52.90 $\pm$ 0.33 
& 39.35 $\pm$ 0.72 
& 24.71 $\pm$ 1.03 
& 40.19 $\pm$ 0.85 \\

\midrule
NeuroOnline.a 
& 53.71 $\pm$ 0.29 
& 47.48 $\pm$ 0.31 
& 53.56 $\pm$ 0.30 
& 44.87 $\pm$ 1.38 
& 32.44 $\pm$ 1.87 
& 46.36 $\pm$ 1.48 \\

\midrule
\rowcolor{blue!15}
\textbf{NeuroOnline***} 
& \textbf{54.03} $\pm$ 1.65
& \textbf{47.78} $\pm$ 1.81
& \textbf{53.76} $\pm$ 1.55
& \textbf{59.37} $\pm$ 0.56
& \textbf{50.35} $\pm$ 0.67
& \textbf{60.56} $\pm$ 0.56 \\

\rowcolor{blue!15}
$\Delta IMP \uparrow$
& \textbf{+0.60\%} 
& \textbf{+0.63\%} 
& \textbf{+0.37\%} 
& \textbf{+32.32\%} 
& \textbf{+55.21\%} 
& \textbf{+30.63\%} \\

\toprule
& \multicolumn{6}{c}{\textbf{LaBraM}} \\
\cmidrule(lr){1-7}
& \multicolumn{3}{c}{\textbf{FACED, 9-class}} 
& \multicolumn{3}{c}{\textbf{SEED-V, 5-class}} \\
\cmidrule(lr){2-4} \cmidrule(lr){5-7}
\textbf{Methods}
& \textbf{Bal.Acc.}
& \textbf{Kappa} 
& \textbf{F1-w}
& \textbf{Bal.Acc.}
& \textbf{Kappa} 
& \textbf{F1-w} \\

\midrule
off-LoRA-LIN 
& 37.49 $\pm$ 0.71 
& 29.32 $\pm$ 0.72 
& 36.95 $\pm$ 0.52 
& 39.90 $\pm$ 0.39 
& 24.75 $\pm$ 0.57 
& 40.17 $\pm$ 0.50 \\

off-LoRA-MLP 
& 46.27 $\pm$ 1.13 
& 39.34 $\pm$ 1.33 
& 46.40 $\pm$ 1.12 
& 41.34 $\pm$ 0.74 
& 26.82 $\pm$ 1.01 
& 41.80 $\pm$ 0.84 \\

off-FT-LIN 
& 42.52 $\pm$ 0.88 
& 35.04 $\pm$ 1.08
& 42.28 $\pm$ 0.99 
& 39.29 $\pm$ 1.16 
& 24.15 $\pm$ 1.51 
& 39.64 $\pm$ 1.39 \\

off-FT-MLP 
& 53.64 $\pm$ 0.42 
& 47.75 $\pm$ 0.49 
& 54.21 $\pm$ 0.48 
& 41.24 $\pm$ 1.02 
& 26.79 $\pm$ 1.48 
& 41.63 $\pm$ 1.36 \\

\midrule
NeuroOnline.a 
& 54.95 $\pm$ 0.30 
& 48.97 $\pm$ 0.32 
& 55.04 $\pm$ 0.32
& 41.27 $\pm$ 0.50 
& 27.95 $\pm$ 0.71 
& 42.70 $\pm$ 0.55 \\

\midrule
\rowcolor{blue!15}
\textbf{NeuroOnline***} 
& \textbf{56.06} $\pm$ 0.82
& \textbf{50.10} $\pm$ 0.90
& \textbf{55.92} $\pm$ 0.79
& \textbf{53.09} $\pm$ 0.58
& \textbf{42.49} $\pm$ 0.68
& \textbf{54.29} $\pm$ 0.55 \\

\rowcolor{blue!15}
$\Delta IMP \uparrow$
& \textbf{+2.02\%} 
& \textbf{+2.31\%} 
& \textbf{+1.60\%} 
& \textbf{+28.42\%} 
& \textbf{+52.02\%} 
& \textbf{+27.14\%} \\

\bottomrule
\end{tabular}
}
\end{table*}



\textbf{Effectiveness Across Additional EEG Foundation Models:}
To further validate the applicability of our approach, we extend the evaluation to additional EEG foundation models, including EEGPT \cite{NEURIPS2024_4540d267} and BIOT \cite{NEURIPS2023_f6b30f3e}, on representative datasets (BCIC-2A and PhysioNet-MI for motor imagery, FACED and SEED-V for emotion recognition). As shown in Table \ref{tab:additional_EEG_foundation_result}, our method consistently outperforms the corresponding baselines across different models and evaluation metrics, with statistical significance assessed against the strongest baseline using the Wilcoxon signed-rank test ($p < 0.001$). NeuroOnline achieves an average relative improvement of 16.29\% (12.85\% for Bal.Acc., 23.49\% for Kappa, and 12.54\% for F1-w) across different datasets and backbone models. These results indicate that the improvements are not tied to backbone-specific designs and demonstrate the broad applicability of NeuroOnline across diverse EEG foundation models.

\begin{table*}[h]
\centering
\caption{Performance comparison across additional EEG foundation models (EEGPT and BIOT) under online settings.}
\label{tab:additional_EEG_foundation_result}
\resizebox{\textwidth}{!}{
\begin{tabular}{lcccccc}
\toprule
& \multicolumn{6}{c}{\textbf{EEGPT}} \\
\cmidrule(lr){1-7}
& \multicolumn{3}{c}{\textbf{BCIC-2A, 4-class}} 
& \multicolumn{3}{c}{\textbf{FACED, 9-class}} \\
\cmidrule(lr){2-4} \cmidrule(lr){5-7}
\textbf{Methods}
& \textbf{Bal.Acc.}
& \textbf{Kappa} 
& \textbf{F1-w}
& \textbf{Bal.Acc.}
& \textbf{Kappa} 
& \textbf{F1-w} \\

\midrule
off-LoRA-LIN 
& 37.20 $\pm$ 2.10
& 16.27 $\pm$ 2.80
& 32.12 $\pm$ 3.62
& 28.39 $\pm$ 0.65
& 19.42 $\pm$ 0.69
& 28.12 $\pm$ 0.73 \\

off-LoRA-MLP 
& 37.31 $\pm$ 1.02
& 16.41 $\pm$ 1.36
& 30.71 $\pm$ 1.74
& 33.60 $\pm$ 1.06
& 25.20 $\pm$ 1.20
& 33.51 $\pm$ 1.14 \\

off-FT-LIN 
& 37.07 $\pm$ 2.19
& 16.09 $\pm$ 2.92
& 32.47 $\pm$ 3.07
& 33.70 $\pm$ 1.01
& 25.34 $\pm$ 1.11
& 33.58 $\pm$ 1.14 \\

off-FT-MLP 
& 37.88 $\pm$ 2.15
& 17.18 $\pm$ 2.87
& 33.35 $\pm$ 2.62
& 39.60 $\pm$ 2.25
& 32.08 $\pm$ 2.45
& 39.96 $\pm$ 2.17 \\

\midrule
NeuroOnline.a 
& 50.99 $\pm$ 1.35
& 34.65 $\pm$ 1.81
& 50.61 $\pm$ 1.52
& 43.08 $\pm$ 1.61
& 35.70 $\pm$ 1.78
& 43.04 $\pm$ 1.53 \\

\midrule
\rowcolor{blue!15}
\textbf{NeuroOnline***} 
& \textbf{53.35} $\pm$ 2.06
& \textbf{37.80} $\pm$ 2.74
& \textbf{52.46} $\pm$ 2.09
& \textbf{49.28} $\pm$ 0.70
& \textbf{42.59} $\pm$ 0.76
& \textbf{49.39} $\pm$ 0.64 \\

\rowcolor{blue!15}
$\Delta IMP \uparrow$
& \textbf{+4.63\%} 
& \textbf{+9.09\%} 
& \textbf{+3.66\%} 
& \textbf{+14.39\%} 
& \textbf{+19.30\%} 
& \textbf{+14.75\%} \\

\toprule
& \multicolumn{6}{c}{\textbf{BIOT}} \\
\cmidrule(lr){1-7}
& \multicolumn{3}{c}{\textbf{Physionet-MI, 4-class}} 
& \multicolumn{3}{c}{\textbf{SEED-V, 5-class}} \\
\cmidrule(lr){2-4} \cmidrule(lr){5-7}
\textbf{Methods}
& \textbf{Bal.Acc.}
& \textbf{Kappa} 
& \textbf{F1-w}
& \textbf{Bal.Acc.}
& \textbf{Kappa} 
& \textbf{F1-w} \\

\midrule
off-LoRA-LIN
& 38.56 $\pm$ 3.23
& 18.07 $\pm$ 4.30
& 37.79 $\pm$ 3.44
& 34.13 $\pm$ 0.62
& 17.79 $\pm$ 0.73
& 34.74 $\pm$ 0.70 \\

off-LoRA-MLP
& 36.88 $\pm$ 3.04
& 15.83 $\pm$ 4.05
& 36.28 $\pm$ 3.16
& 34.39 $\pm$ 1.41
& 17.91 $\pm$ 1.64
& 34.83 $\pm$ 1.36 \\

off-FT-LIN
& 45.38 $\pm$ 3.41
& 27.17 $\pm$ 4.55
& 45.25 $\pm$ 3.43
& 35.47 $\pm$ 1.90
& 19.40 $\pm$ 2.29
& 35.98 $\pm$ 2.07 \\

off-FT-MLP
& 43.79 $\pm$ 4.06
& 25.04 $\pm$ 5.41
& 43.60 $\pm$ 4.02
& 36.17 $\pm$ 1.47
& 20.27 $\pm$ 1.74
& 36.80 $\pm$ 1.50 \\

\midrule
NeuroOnline.a 
& 42.82 $\pm$ 2.10
& 23.77 $\pm$ 2.81
& 42.61 $\pm$ 2.08
& 38.28 $\pm$ 1.02
& 24.30 $\pm$ 1.26
& 39.76 $\pm$ 0.99 \\

\midrule
\rowcolor{blue!15}
\textbf{NeuroOnline***} 
& \textbf{49.52} $\pm$ 3.30
& \textbf{32.69} $\pm$ 4.41
& \textbf{49.72} $\pm$ 3.33
& \textbf{47.18} $\pm$ 0.77
& \textbf{35.29} $\pm$ 1.03
& \textbf{48.45} $\pm$ 0.83 \\

\rowcolor{blue!15}
$\Delta IMP \uparrow$
& \textbf{+9.12\%} 
& \textbf{+20.32\%} 
& \textbf{+9.88\%} 
& \textbf{+23.25\%} 
& \textbf{+45.23\%} 
& \textbf{+21.86\%} \\

\bottomrule
\end{tabular}
}
\end{table*}



\textbf{Analysis:} We draw two key observations from the experimental results: (1) Compared to offline methods, the proposed approach exhibits stronger adaptability under distribution shifts. Compared to online updating baselines, it not only achieves better performance but also demonstrates improved stability, suggesting that naive parameter updating strategies are insufficient for handling evolving data distributions in dynamic environments. (2) Across different downstream tasks and backbone architectures, our method consistently achieves strong performance, demonstrating its effectiveness and broad applicability across models and tasks. These improvements can be attributed to two key factors: (1) Multi-view consistency learning enforces alignment across diverse views of online data, promoting more consistent and task-relevant representations. (2) The context-aware modulation mechanism dynamically adjusts representations based on input-dependent context, enabling effective adaptation under evolving data distributions. Together, these components unify representation alignment and dynamic adaptation, leading to consistent improvements in performance under distribution shifts. This highlights the importance of jointly modeling representation consistency and context-dependent adaptation in online learning settings.

\begin{table*}[t]
\centering
\caption{Ablation experiment.}
\label{tab:ablation_experiment}
\resizebox{\textwidth}{!}{
\begin{tabular}{lcccccc}
\toprule
& \multicolumn{6}{c}{\textbf{CBraMod}} \\
\cmidrule(lr){1-7}
& \multicolumn{3}{c}{\textbf{BCIC-2A, 4-class}} 
& \multicolumn{3}{c}{\textbf{BCIC-2B, 2-class}} \\
\cmidrule(lr){2-4} \cmidrule(lr){5-7}
\textbf{Methods}
& \textbf{Bal.Acc.}
& \textbf{Kappa} 
& \textbf{F1-w}
& \textbf{Bal.Acc.}
& \textbf{AUC-PR} 
& \textbf{AUROC} \\

\midrule
w/o Consistency
& 62.71 $\pm$ 1.04 
& 50.28 $\pm$ 1.39 
& 62.47 $\pm$ 1.17 
& 65.50 $\pm$ 0.77 
& 71.60 $\pm$ 0.59 
& 71.48 $\pm$ 0.45  \\

w/o Prompt
& 64.57 $\pm$ 2.26 
& 52.75 $\pm$ 3.01 
& 64.38 $\pm$ 2.26 
& 66.57 $\pm$ 0.99 
& 72.55 $\pm$ 1.54 
& 72.85 $\pm$ 1.35  \\

w/o CRM
& 62.95 $\pm$ 1.03 
& 50.60 $\pm$ 1.38 
& 62.76 $\pm$ 1.07 
& 65.73 $\pm$ 0.58 
& 72.11 $\pm$ 0.67 
& 72.20 $\pm$ 0.58  \\

\midrule
\rowcolor{blue!15}
\textbf{NeuroOnline}
& \textbf{66.18} $\pm$ 2.11
& \textbf{54.91} $\pm$ 2.81
& \textbf{66.02} $\pm$ 2.11
& \textbf{66.74} $\pm$ 0.61
& \textbf{72.63} $\pm$ 1.55
& \textbf{73.04} $\pm$ 1.29 \\

\midrule
& \multicolumn{6}{c}{\textbf{LaBraM}} \\
\cmidrule(lr){1-7}
& \multicolumn{3}{c}{\textbf{BCIC-2A, 4-class}} 
& \multicolumn{3}{c}{\textbf{SHU-MI, 2-class}} \\
\cmidrule(lr){2-4} \cmidrule(lr){5-7}
\textbf{Methods}
& \textbf{Bal.Acc.}
& \textbf{Kappa} 
& \textbf{F1-w}
& \textbf{Bal.Acc.}
& \textbf{AUC-PR} 
& \textbf{AUROC} \\

\midrule
w/o Consistency
& 63.89 $\pm$ 1.86 
& 51.85 $\pm$ 2.49 
& 63.77 $\pm$ 1.98 
& 63.06 $\pm$ 0.37 
& 69.54 $\pm$ 0.86 
& 69.11 $\pm$ 0.75  \\

w/o Prompt
& 63.51 $\pm$ 1.38 
& 51.34 $\pm$ 1.84 
& 63.23 $\pm$ 1.44 
& 63.79 $\pm$ 0.88 
& 69.26 $\pm$ 1.25 
& 69.27 $\pm$ 1.11  \\

w/o CRM
& 64.43 $\pm$ 1.08 
& 52.57 $\pm$ 1.44 
& 64.17 $\pm$ 1.09 
& 63.38 $\pm$ 0.53
& 69.41 $\pm$ 0.97 
& 69.80 $\pm$ 1.23  \\

\midrule
\rowcolor{blue!15}
\textbf{NeuroOnline}
& \textbf{66.18} $\pm$ 0.85
& \textbf{54.91} $\pm$ 1.13
& \textbf{66.03} $\pm$ 0.86
& \textbf{63.92} $\pm$ 0.52
& \textbf{70.17} $\pm$ 0.87
& \textbf{70.16} $\pm$ 0.50 \\

\bottomrule
\end{tabular}
}
\end{table*}

\subsection{Ablation Study}
To validate the effectiveness of each component in the proposed framework, we conduct systematic ablation studies. All experiments are conducted under the same online setting as the main experiments for fair comparison.




\textbf{Setup:} We construct three model variants: (1) without multi-view consistency (w/o Consistency); (2) without the Context Prompt (w/o Prompt); and (3) without the context-aware representation modulation module (w/o CRM). \textbf{Results and Analysis:} As shown in Table \ref{tab:ablation_experiment}, removing each key component leads to performance degradation, confirming the necessity of each module. Further analysis shows that these components contribute in a complementary manner. Specifically, the multi-view consistency constraint enforces cross-view representation alignment, improving task consistency; the context prompt introduces input-dependent contextual information to capture underlying distributional variations; and CRM enables dynamic modulation of representations, allowing the model to adapt to evolving data distributions. Overall, these components operate on representation alignment, context modeling, and dynamic adaptation, respectively, and their synergy is essential for achieving stable and sustained performance in non-stationary environments.

\subsection{Sensitivity Analysis}
To evaluate the sensitivity of the proposed method to key hyperparameters, we analyze the effect of the consistency loss weight $\lambda$ on model performance. \textbf{Results:} As shown in Table \ref{tab:hyperparameter_experiment}, performance first improves and then degrades as $\lambda$ increases. When $\lambda$ is small, the consistency constraint is too weak to effectively enhance representation alignment. Conversely, overly large $\lambda$ suppresses the classification objective, leading to performance decline. Notably, the model remains stable across a relatively wide range of $\lambda$. \textbf{Analysis:} These results demonstrate that the proposed method is robust to variations in the key hyperparameter and does not require careful tuning to achieve strong performance.

\begin{table*}[t]
\centering
\caption{Hyperparameter analysis.}
\label{tab:hyperparameter_experiment}
\resizebox{\textwidth}{!}{
\begin{tabular}{lcccccc}
\toprule
& \multicolumn{6}{c}{\textbf{CBraMod}} \\
\cmidrule(lr){1-7}
& \multicolumn{3}{c}{\textbf{BCIC-2A, 4-class}}
& \multicolumn{3}{c}{\textbf{BCIC-2B, 2-class}} \\
\cmidrule(lr){2-4}
\cmidrule(lr){2-4} \cmidrule(lr){5-7}
\textbf{$\lambda$}
& \textbf{Bal.Acc.}
& \textbf{Kappa} 
& \textbf{F1-w}
& \textbf{Bal.Acc.}
& \textbf{AUC-PR} 
& \textbf{AUROC} \\

\midrule
0.10
& 66.09 $\pm$ 2.24 
& 54.79 $\pm$ 2.99 
& 65.94 $\pm$ 2.24
& 66.32 $\pm$ 1.41
& 72.01 $\pm$ 1.42 
& 72.49 $\pm$ 1.56  \\

0.25
& 66.15 $\pm$ 1.94 
& 54.86 $\pm$ 2.59 
& 65.98 $\pm$ 1.95
& 66.45 $\pm$ 1.00
& 72.41 $\pm$ 1.56 
& 72.81 $\pm$ 1.45  \\

0.50
& \cellcolor{blue!15} \textbf{66.18} $\pm$ 2.11 
& \cellcolor{blue!15} \textbf{54.91} $\pm$ 2.81 
& \cellcolor{blue!15} \textbf{66.02} $\pm$ 2.11
& 66.72 $\pm$ 1.00
& 72.60 $\pm$ 1.33 
& 73.01 $\pm$ 1.31  \\

1.00
& 65.94 $\pm$ 1.87 
& 54.58 $\pm$ 2.49 
& 65.76 $\pm$ 1.84
& \cellcolor{blue!15} \textbf{66.74} $\pm$ 0.61
& \cellcolor{blue!15} \textbf{72.63} $\pm$ 1.55 
& \cellcolor{blue!15} \textbf{73.04} $\pm$ 1.29  \\

2.00
& 65.82 $\pm$ 1.78 
& 54.42 $\pm$ 2.38 
& 65.65 $\pm$ 1.78
& 66.53 $\pm$ 0.72
& 72.53 $\pm$ 1.23 
& 72.98 $\pm$ 1.12  \\

\midrule
& \multicolumn{6}{c}{\textbf{LaBraM}} \\
\cmidrule(lr){1-7}
& \multicolumn{3}{c}{\textbf{BCIC-2A, 4-class}}
& \multicolumn{3}{c}{\textbf{PhysioNet-MI, 4-class}} \\
\cmidrule(lr){2-4}
\cmidrule(lr){2-4} \cmidrule(lr){5-7}
\textbf{$\lambda$}
& \textbf{Bal.Acc.}
& \textbf{Kappa} 
& \textbf{F1-w}
& \textbf{Bal.Acc.}
& \textbf{Kappa} 
& \textbf{F1-w} \\

\midrule
0.10
& 65.89 $\pm$ 0.96
& 54.51 $\pm$ 1.28 
& 65.74 $\pm$ 0.95
& 60.51 $\pm$ 0.83
& 47.36 $\pm$ 1.12 
& 60.48 $\pm$ 1.04  \\

0.25
& \cellcolor{blue!15} \textbf{66.18} $\pm$ 0.85
& \cellcolor{blue!15} \textbf{54.91} $\pm$ 1.13 
& \cellcolor{blue!15} \textbf{66.03} $\pm$ 0.86
& \cellcolor{blue!15} \textbf{60.53} $\pm$ 0.80
& \cellcolor{blue!15} \textbf{47.38} $\pm$ 1.06 
& \cellcolor{blue!15} \textbf{60.50} $\pm$ 0.94  \\

0.50
& 65.92 $\pm$ 1.14
& 54.56 $\pm$ 1.51 
& 65.75 $\pm$ 1.16
& 60.50 $\pm$ 0.78
& 47.34 $\pm$ 1.04 
& 60.47 $\pm$ 0.92  \\

1.00
& 65.75 $\pm$ 1.01
& 54.33 $\pm$ 1.35 
& 65.55 $\pm$ 1.04
& 60.52 $\pm$ 0.79
& 47.36 $\pm$ 1.06 
& 60.49 $\pm$ 0.94  \\

2.00
& 65.56 $\pm$ 1.27
& 54.07 $\pm$ 1.69 
& 65.33 $\pm$ 1.32
& 60.50 $\pm$ 0.81
& 47.33 $\pm$ 1.08 
& 60.47 $\pm$ 0.96  \\

\bottomrule
\end{tabular}
}
\end{table*}

\section{Conclusion}
In this work, we propose NeuroOnline, a unified framework for continuous adaptation of EEG foundation models in online environments. It integrates multi-view consistency learning and context-aware representation modulation to jointly model representation alignment and dynamic adaptation. Extensive experiments across multiple EEG benchmarks demonstrate that NeuroOnline consistently outperforms strong baselines in online settings. Ablation and sensitivity analyses further validate the effectiveness of the overall design. These results highlight the importance of jointly addressing representation alignment and dynamic adaptation for EEG foundation models in non-stationary environments. \textbf{Limitations and Future Work:} The proposed method is primarily empirical and lacks rigorous theoretical analysis, which we leave for future work. Although the proposed method performs well across multiple benchmarks, its scalability to larger settings remains to be systematically evaluated. Future work will explore more expressive context modeling strategies and more efficient online adaptation mechanisms to further enhance generalization in real-world scenarios.

\bibliographystyle{unsrtnat}
\bibliography{references}








\appendix

\section{Algorithm}
\label{Algorithm}
The overall optimization procedure of NeuroOnline is summarized in Algorithm~\ref{algorithm}.

\begin{algorithm}
\caption{The Pipeline of NeuroOnline}
\label{algorithm} 
\textbf{(1) Online Training Phase:}\\
\textbf{Input:} Observed samples.\\
\textbf{Step 1: Multi-view construction}\\
For each sample $X$, treat it as the anchor view $X^{(0)}$ and generate $K$ augmented views $\{X^{(i)}\}_{i=1}^{K}$ via stochastic transformations, including temporal and frequency masking.\\
\textbf{Step 2: Context-aware representation modulation}\\
Compute the modulated representations for the anchor and augmented views via Eq.\eqref{modulation}: $\tilde{z}^{(i)}=g_\theta(f_\theta(X^{(i)}),\mathcal{P}),\ i=0,\dots,K$.\\
\textbf{Step 3: Multi-view consistency learning}\\
Compute consistency loss $\mathcal{L}_{\text{cons}}$ via Eq.\eqref{lcons},\\
\textbf{Step 4: Online optimization}\\
Compute overall loss $\mathcal{L} = \mathcal{L}_{\text{cls}} + \lambda \mathcal{L}_{\text{cons}}$ via Eq.\eqref{total-loss},\\
Update model parameters by minimizing $\mathcal{L}$.\\
\textbf{(2) Online Inference Phase:}\\
\textbf{Input:} Current EEG sample $X_t$,\\
Predict label $\hat{y}_t$ with fixed model parameters.
\end{algorithm}

\section{Additional Details on Datasets and Preprocessing}
\label{Additional Details on Datasets and Preprocessing}

To better reflect real-world online settings, preprocessing is performed on individual EEG segments (i.e., epochs or trials) rather than on entire continuous EEG recordings.

\textbf{Motor Imagery Classification.}

BCIC-2A \cite{brunner2008bci} is a motor imagery (MI) dataset containing EEG recordings from 9 subjects using 22 channels. It includes four classes: left hand, right hand, both feet, and tongue movements. The continuous EEG signals are segmented into 4-second segments (i.e., epochs), resulting in 5088 samples. Each segment is band-pass filtered between 0.3–50 Hz and resampled to 200 Hz to remove low- and high-frequency noise. We adopt a strict cross-subject evaluation protocol, where subjects 1–5 are used for training, 6–7 for validation, and 8–9 for testing.

BCIC-2B \cite{steyrl2016random} is a motor imagery dataset consisting of EEG recordings from 9 subjects using 3 channels, with two classes: left- and right-hand motor imagery. The EEG signals are segmented into 4-second segments (i.e., epochs), yielding 6520 samples. Each epoch is band-pass filtered between 0–38 Hz and resampled to 200 Hz. To reduce artifacts, we remove electrooculogram (EOG) channels and apply common average referencing (CAR) across channels. We use the same cross-subject protocol as above, with subjects 1–5 for training, 6–7 for validation, and 8–9 for testing.

SHU-MI \cite{ma2022large} is a motor imagery (MI) EEG dataset containing recordings from 25 subjects with 32 channels, covering two classes: left- and right-hand imagery. The EEG signals are segmented into 4-second segments (i.e., epochs), yielding 11,988 samples, and are resampled to 200 Hz. We adopt a strict cross-subject protocol, using subjects 1–15 for training, 16–20 for validation, and 21–25 for testing.


PhysioNet-MI \cite{goldberger2000physiobank, schalk2004bci2000} is a motor imagery (MI) EEG dataset containing recordings from 109 subjects with 64 channels, covering four classes: left fist, right fist, both fists, and both feet movements. The EEG signals are segmented into 4-second segments (i.e., epochs), yielding 9,837 samples. We apply common average referencing (CAR), followed by a 0.3 Hz high-pass filter to remove low-frequency drift. A 60 Hz notch filter is used to suppress power-line noise, and the signals are then resampled to 200 Hz. We adopt a strict cross-subject protocol, using subjects 1–70 for training, 71–89 for validation, and 90–109 for testing.

\textbf{ Emotion Recognition.}


FACED \cite{chen2023large} is a large-scale fine-grained affective EEG dataset containing recordings from 123 subjects with 32 channels, covering nine emotion classes: amusement, inspiration, joy, tenderness, anger, fear, disgust, sadness, and neutral. The EEG signals are segmented into 10-second segments, yielding 10,332 samples, and are resampled to 200 Hz. We adopt a cross-subject protocol, using subjects 1–80 for training, 81–100 for validation, and 101–123 for testing.


SEED-V \cite{liu2021comparing} is an affective EEG dataset containing recordings from 16 subjects with 62 channels, covering five emotion classes: happiness, sadness, neutral, disgust, and fear. Each subject has three sessions, and each session consists of 15 trials. According to the official documentation, one session is corrupted and is therefore excluded from our experiments. During preprocessing, we first remove irrelevant channels (M1, M2, VEO, and HEO) and extract trials based on the provided onset and offset annotations. The EEG signals are then resampled to 200 Hz and band-pass filtered between 0.3–75 Hz to remove low-frequency drift and high-frequency noise. Each trial is further segmented into non-overlapping 1-second segments, resulting in 115,001 samples. Within each session, the 15 trials are evenly divided into three parts (5:5:5) for training, validation, and testing, respectively.

\section{Additional Details on Baselines}
\label{Additional Details on Baselines}

\textbf{(1) Offline baselines.} 

Offline baselines are fine-tuned on the training set and directly evaluated on the test set without any online updates. To ensure fair comparison, all methods share the same pretrained backbone and follow identical data splits and optimization settings. We consider two categories of fine-tuning strategies, i.e., parameter-efficient adaptation (LoRA \cite{hu2022lora}) and full fine-tuning (FT), each with either a linear or an MLP classification head:

(i) \textbf{off-LoRA-LIN:} We inject LoRA modules into the backbone and optimize only the low-rank parameters together with a linear classification head, while keeping the original backbone weights frozen;


(ii) \textbf{off-LoRA-MLP:} We adopt the same LoRA setting as in (i) but replace the linear head with an MLP classifier. Following CBraMod \cite{wang2024cbramod}, the backbone output is flattened and passed through two hidden layers, each followed by ELU activation and dropout, before a final linear layer for prediction;

(iii) \textbf{off-FT-LIN:} We use a single linear classification head and jointly optimize the backbone and the classifier, serving as a lightweight full fine-tuning baseline;

(iv) \textbf{off-FT-MLP:} We attach the same MLP classification head as in (ii) and fine-tune the entire model end-to-end.

For LoRA-based methods, we uniformly apply LoRA modules to all linear layers in the backbone (including attention projections and feed-forward layers, except the output projection), enabling efficient adaptation without significantly increasing the number of parameters.

All offline methods are fixed after training and directly used for inference at test time, serving as baselines without online adaptability.

\textbf{(2) Online baselines.}

\textbf{NeuroOnline.a:} In the online setting, the model performs online adaptation under the same classification objective as in offline training. To prevent information leakage, only online observed samples are used for updates. NeuroOnline.a is a simplified variant of our method, where the MCL and CRM modules are removed, reducing it to a standard online fine-tuning strategy. This serves as a practical online baseline given the lack of established methods for EEG foundation models.

\section{More Details for Backbone Models}
\label{More Details for Backbone Models}

CBraMod \cite{wang2024cbramod} adopts a criss-cross transformer with parallel spatial–temporal attention and masked pretraining to learn scalable and generalizable EEG representations.

LaBraM \cite{ICLR2024_47393e85} learns generalizable EEG representations via vector-quantized tokenization and masked Transformer pre-training on large-scale unlabeled data.

EEGPT \cite{NEURIPS2024_4540d267} employs a mask-based dual self-supervised objective with spatio-temporal representation alignment to learn high-quality and transferable EEG representations from low-SNR signals.

BIOT \cite{NEURIPS2023_f6b30f3e} addresses heterogeneous biosignal formats by tokenizing multi-channel signals into a unified representation and modeling them with transformers for cross-dataset learning.

\section{Details of Evaluation Metrics}

\label{Details of Evaluation Metrics}

To comprehensively evaluate model performance, following CBraMod \cite{wang2024cbramod}, we adopt several widely used metrics for downstream tasks.

\textbf{Balanced Accuracy.} Balanced Accuracy computes the average recall across classes, thereby mitigating the impact of class imbalance. It is used for both binary and multi-class classification.

\textbf{AUC-PR.} The area under the precision–recall curve (AUC-PR) reflects the trade-off between precision and recall across different thresholds and is particularly suitable for imbalanced binary classification.

\textbf{AUROC.} The area under the receiver operating characteristic curve (AUROC) measures the model’s ability to distinguish between classes across varying decision thresholds. It is commonly used for evaluating binary classification performance.

\textbf{Cohen’s Kappa.} Cohen’s Kappa measures the agreement between predicted labels and ground truth while accounting for chance agreement. It naturally extends to multi-class settings and provides a more robust evaluation than accuracy, particularly when class distributions are imbalanced.

\textbf{Weighted F1.} Weighted F1 computes the average F1-score across classes, where each class is weighted by its sample proportion, providing a more comprehensive evaluation in multi-class classification.

\section{The Details of hyperparameter settings}
\label{The Details of hyperparameter settings}

According to CBraMod \cite{wang2024cbramod}, the hyperparameter configurations for the offline baselines are provided in Table \ref{tab:hyperparameters for offline baselines}, while those for the online baselines and our method are reported in Table \ref{tab:hyperparameters for online baselines and ours}.

\begin{table}[t]
\centering
\caption{Hyperparameter settings for offline baselines.}
\label{tab:hyperparameters for offline baselines}
\begin{tabular}{ll}
\toprule
\textbf{Hyperparameters} & \textbf{Settings} \\
\midrule
Epochs & 50 \\
Batch size & 64 \\
Dropout & 0.1 \\
Optimizer & AdamW \\
Learning rate & 1e-4 \\
Weight decay & 5e-2 \\
Scheduler & CosineAnnealingLR \\
Cosine cycle epochs & 50 \\
Minimal learning rate & 1e-6 \\
Clipping gradient norm & 1 \\
Label smoothing (multi-class classification) & 0.1 \\
\bottomrule
\end{tabular}
\end{table}

\begin{table}[t]
\centering
\caption{Hyperparameter settings for online baselines and our method.}
\label{tab:hyperparameters for online baselines and ours}
\begin{tabular}{ll}
\toprule
\textbf{Hyperparameters} & \textbf{Settings} \\
\midrule
Epochs & 3 \\
Batch size & 64 \\
Dropout & 0.1 \\
Optimizer & AdamW \\
Learning rate & 1e-4 \\
Weight decay & 5e-2 \\
Label smoothing (multi-class classification) & 0.1 \\
\bottomrule
\end{tabular}
\end{table}




\newpage
\input{checklist.tex}

\end{document}

%% file: checklist.tex
\section*{NeurIPS Paper Checklist}

\begin{enumerate}

\item {\bf Claims}
    \item[] Question: Do the main claims made in the abstract and introduction accurately reflect the paper's contributions and scope?
    \item[] Answer: \answerYes{} 
    \item[] Justification: The claims in the abstract and introduction accurately reflect the paper’s contributions. Specifically, the paper introduces the NeuroOnline framework for bridging pretraining and online adaptation of EEG foundation models, and proposes two key mechanisms: multi-view consistency learning for representation alignment and context-aware modulation for handling distribution shifts. These claims are supported by extensive experiments on multiple EEG benchmarks, demonstrating consistent improvements in online settings.

\item {\bf Limitations}
    \item[] Question: Does the paper discuss the limitations of the work performed by the authors?
    \item[] Answer: \answerYes{} 
    \item[] Justification: Yes, the limitations have been discussed in Section 5.

\item {\bf Theory assumptions and proofs}
    \item[] Question: For each theoretical result, does the paper provide the full set of assumptions and a complete (and correct) proof?
    \item[] Answer: \answerNA{} 
    \item[] Justification: This work is focused on experiments and engineering, and does not include results that require theoretical proof.

    \item {\bf Experimental result reproducibility}
    \item[] Question: Does the paper fully disclose all the information needed to reproduce the main experimental results of the paper to the extent that it affects the main claims and/or conclusions of the paper (regardless of whether the code and data are provided or not)?
    \item[] Answer: \answerYes{} 
    \item[] Justification: This paper provides all the necessary details to reproduce the main results.

\item {\bf Open access to data and code}
    \item[] Question: Does the paper provide open access to the data and code, with sufficient instructions to faithfully reproduce the main experimental results, as described in supplemental material?
    \item[] Answer: \answerYes{} 
    \item[] Justification: We provide all code, data preprocessing scripts, and detailed instructions in the supplemental material to reproduce the main experimental results.

\item {\bf Experimental setting/details}
    \item[] Question: Does the paper specify all the training and test details (e.g., data splits, hyperparameters, how they were chosen, type of optimizer) necessary to understand the results?
    \item[] Answer: \answerYes{} 
    \item[] Justification: The paper specifies all necessary training and test details, including data splits, hyperparameters, optimizer settings, and training procedures, in Section 4 and Appendix A–F.

\item {\bf Experiment statistical significance}
    \item[] Question: Does the paper report error bars suitably and correctly defined or other appropriate information about the statistical significance of the experiments?
    \item[] Answer: \answerYes{} 
    \item[] Justification: The results are accompanied by error bars and statistical significance tests.

\item {\bf Experiments compute resources}
    \item[] Question: For each experiment, does the paper provide sufficient information on the computer resources (type of compute workers, memory, time of execution) needed to reproduce the experiments?
    \item[] Answer: \answerYes{} 
    \item[] Justification:  Section 4 provides sufficient information on the computer resources.
    
\item {\bf Code of ethics}
    \item[] Question: Does the research conducted in the paper conform, in every respect, with the NeurIPS Code of Ethics \url{https://neurips.cc/public/EthicsGuidelines}?
    \item[] Answer: \answerYes{} 
    \item[] Justification: All research conducted in this paper conforms fully to the NeurIPS Code of Ethics.

\item {\bf Broader impacts}
    \item[] Question: Does the paper discuss both potential positive societal impacts and negative societal impacts of the work performed?
    \item[] Answer: \answerYes{} 
    \item[] Justification: The paper discusses the potential positive societal impacts, such as advancements in EEG-based diagnostics and therapeutic applications, which could improve healthcare outcomes.
    
\item {\bf Safeguards}
    \item[] Question: Does the paper describe safeguards that have been put in place for responsible release of data or models that have a high risk for misuse (e.g., pre-trained language models, image generators, or scraped datasets)?
    \item[] Answer: \answerNA{} 
    \item[] Justification: The paper poses no such risks.

\item {\bf Licenses for existing assets}
    \item[] Question: Are the creators or original owners of assets (e.g., code, data, models), used in the paper, properly credited and are the license and terms of use explicitly mentioned and properly respected?
    \item[] Answer: \answerYes{} 
    \item[] Justification: Yes, the creators and original owners of all assets used in the paper are properly credited, and the licenses and terms of use are explicitly mentioned and fully respected.

\item {\bf New assets}
    \item[] Question: Are new assets introduced in the paper well documented and is the documentation provided alongside the assets?
    \item[] Answer: \answerYes{} 
    \item[] Justification: Yes, the new assets introduced in the paper are well documented, and the documentation is provided alongside the assets.

\item {\bf Crowdsourcing and research with human subjects}
    \item[] Question: For crowdsourcing experiments and research with human subjects, does the paper include the full text of instructions given to participants and screenshots, if applicable, as well as details about compensation (if any)? 
    \item[] Answer: \answerNA{} 
    \item[] Justification: The paper uses only publicly available datasets and does not involve crowdsourcing nor research with human subjects.

\item {\bf Institutional review board (IRB) approvals or equivalent for research with human subjects}
    \item[] Question: Does the paper describe potential risks incurred by study participants, whether such risks were disclosed to the subjects, and whether Institutional Review Board (IRB) approvals (or an equivalent approval/review based on the requirements of your country or institution) were obtained?
    \item[] Answer: \answerNA{} 
    \item[] Justification:  The paper uses only publicly available datasets and does not involve involve crowdsourcing nor research with human subjects.

\item {\bf Declaration of LLM usage}
    \item[] Question: Does the paper describe the usage of LLMs if it is an important, original, or non-standard component of the core methods in this research? Note that if the LLM is used only for writing, editing, or formatting purposes and does \emph{not} impact the core methodology, scientific rigor, or originality of the research, declaration is not required.
    \item[] Answer: \answerNA{} 
    \item[] Justification: This research does not involve LLMs as any important, original, or non-standard components.

\end{enumerate}

%% file: neurips_2026.bbl
\begin{thebibliography}{41}
\providecommand{\natexlab}[1]{#1}
\providecommand{\url}[1]{\texttt{#1}}
\expandafter\ifx\csname urlstyle\endcsname\relax
  \providecommand{\doi}[1]{doi: #1}\else
  \providecommand{\doi}{doi: \begingroup \urlstyle{rm}\Url}\fi

\bibitem[Abiri et~al.(2019)Abiri, Borhani, Sellers, Jiang, and Zhao]{abiri2019comprehensive}
Reza Abiri, Soheil Borhani, Eric~W Sellers, Yang Jiang, and Xiaopeng Zhao.
\newblock A comprehensive review of eeg-based brain--computer interface paradigms.
\newblock \emph{Journal of neural engineering}, 16\penalty0 (1):\penalty0 011001, 2019.

\bibitem[Aggarwal and Chugh(2022)]{aggarwal2022review}
Swati Aggarwal and Nupur Chugh.
\newblock Review of machine learning techniques for eeg based brain computer interface.
\newblock \emph{Archives of Computational Methods in Engineering}, 29\penalty0 (5):\penalty0 3001--3020, 2022.

\bibitem[Lotte et~al.(2018)Lotte, Bougrain, Cichocki, Clerc, Congedo, Rakotomamonjy, and Yger]{lotte2018review}
Fabien Lotte, Laurent Bougrain, Andrzej Cichocki, Maureen Clerc, Marco Congedo, Alain Rakotomamonjy, and Florian Yger.
\newblock A review of classification algorithms for eeg-based brain--computer interfaces: a 10 year update.
\newblock \emph{Journal of neural engineering}, 15\penalty0 (3):\penalty0 031005, 2018.

\bibitem[V{\"a}rbu et~al.(2022)V{\"a}rbu, Muhammad, and Muhammad]{varbu2022past}
Kaido V{\"a}rbu, Naveed Muhammad, and Yar Muhammad.
\newblock Past, present, and future of eeg-based bci applications.
\newblock \emph{Sensors}, 22\penalty0 (9):\penalty0 3331, 2022.

\bibitem[Min et~al.(2023)Min, Ross, Sulem, Veyseh, Nguyen, Sainz, Agirre, Heintz, and Roth]{10.1145/3605943}
Bonan Min, Hayley Ross, Elior Sulem, Amir Pouran~Ben Veyseh, Thien~Huu Nguyen, Oscar Sainz, Eneko Agirre, Ilana Heintz, and Dan Roth.
\newblock Recent advances in natural language processing via large pre-trained language models: A survey.
\newblock \emph{ACM Comput. Surv.}, 56\penalty0 (2), September 2023.
\newblock ISSN 0360-0300.
\newblock \doi{10.1145/3605943}.
\newblock URL \url{https://doi.org/10.1145/3605943}.

\bibitem[He et~al.(2022)He, Chen, Xie, Li, Dollár, and Girshick]{9879206}
Kaiming He, Xinlei Chen, Saining Xie, Yanghao Li, Piotr Dollár, and Ross Girshick.
\newblock Masked autoencoders are scalable vision learners.
\newblock In \emph{2022 IEEE/CVF Conference on Computer Vision and Pattern Recognition (CVPR)}, pages 15979--15988, 2022.
\newblock \doi{10.1109/CVPR52688.2022.01553}.

\bibitem[Chen et~al.(2020)Chen, Kornblith, Norouzi, and Hinton]{chen2020simple}
Ting Chen, Simon Kornblith, Mohammad Norouzi, and Geoffrey Hinton.
\newblock A simple framework for contrastive learning of visual representations.
\newblock In \emph{International conference on machine learning}, pages 1597--1607. PmLR, 2020.

\bibitem[Devlin et~al.(2019)Devlin, Chang, Lee, and Toutanova]{devlin2019bert}
Jacob Devlin, Ming-Wei Chang, Kenton Lee, and Kristina Toutanova.
\newblock Bert: Pre-training of deep bidirectional transformers for language understanding.
\newblock In \emph{Proceedings of the 2019 conference of the North American chapter of the association for computational linguistics: human language technologies, volume 1 (long and short papers)}, pages 4171--4186, 2019.

\bibitem[Kuruppu et~al.(2026)Kuruppu, Wagh, Kremen, and Varatharajah]{kuruppu2026eeg}
Gayal Kuruppu, Neeraj Wagh, Vaclav Kremen, and Yogatheesan Varatharajah.
\newblock Eeg foundation models: a critical review of current progress and future directions.
\newblock \emph{Journal of neural engineering}, 23\penalty0 (2):\penalty0 021001, 2026.

\bibitem[Lai et~al.(2025)Lai, Wei, Yao, and Wang]{lai2025simple}
Junhong Lai, Jiyu Wei, Lin Yao, and Yueming Wang.
\newblock A simple review of eeg foundation models: Datasets, advancements and future perspectives.
\newblock \emph{arXiv preprint arXiv:2504.20069}, 2025.

\bibitem[Wu et~al.(2025)Wu, Ren, Wang, Zhu, Song, Liu, Zheng, Bai, Ouyang, and Song]{wu2025adabrain}
Jiamin Wu, Zichen Ren, Junyu Wang, Pengyu Zhu, Yonghao Song, Mianxin Liu, Qihao Zheng, Lei Bai, Wanli Ouyang, and Chunfeng Song.
\newblock Adabrain-bench: Benchmarking brain foundation models for brain-computer interface applications.
\newblock \emph{arXiv preprint arXiv:2507.09882}, 2025.

\bibitem[Suhaimi et~al.(2020)Suhaimi, Mountstephens, and Teo]{https://doi.org/10.1155/2020/8875426}
Nazmi~Sofian Suhaimi, James Mountstephens, and Jason Teo.
\newblock Eeg-based emotion recognition: A state-of-the-art review of current trends and opportunities.
\newblock \emph{Computational Intelligence and Neuroscience}, 2020\penalty0 (1):\penalty0 8875426, 2020.
\newblock \doi{https://doi.org/10.1155/2020/8875426}.
\newblock URL \url{https://onlinelibrary.wiley.com/doi/abs/10.1155/2020/8875426}.

\bibitem[Boonyakitanont et~al.(2020)Boonyakitanont, Lek-uthai, Chomtho, and Songsiri]{BOONYAKITANONT2020101702}
Poomipat Boonyakitanont, Apiwat Lek-uthai, Krisnachai Chomtho, and Jitkomut Songsiri.
\newblock A review of feature extraction and performance evaluation in epileptic seizure detection using eeg.
\newblock \emph{Biomedical Signal Processing and Control}, 57:\penalty0 101702, 2020.
\newblock ISSN 1746-8094.
\newblock \doi{https://doi.org/10.1016/j.bspc.2019.101702}.
\newblock URL \url{https://www.sciencedirect.com/science/article/pii/S1746809419302836}.

\bibitem[Sharma et~al.(2022)Sharma, Bohat, Habib, Al-Zoubi, Faris, and Aljarah]{SHARMA2022116634}
Lakhan~Dev Sharma, Vijay~Kumar Bohat, Maria Habib, Ala’~M. Al-Zoubi, Hossam Faris, and Ibrahim Aljarah.
\newblock Evolutionary inspired approach for mental stress detection using eeg signal.
\newblock \emph{Expert Systems with Applications}, 197:\penalty0 116634, 2022.
\newblock ISSN 0957-4174.
\newblock \doi{https://doi.org/10.1016/j.eswa.2022.116634}.
\newblock URL \url{https://www.sciencedirect.com/science/article/pii/S0957417422001233}.

\bibitem[Amin et~al.(2019)Amin, Alsulaiman, Muhammad, Mekhtiche, and {Shamim Hossain}]{AMIN2019542}
Syed~Umar Amin, Mansour Alsulaiman, Ghulam Muhammad, Mohamed~Amine Mekhtiche, and M.~{Shamim Hossain}.
\newblock Deep learning for eeg motor imagery classification based on multi-layer cnns feature fusion.
\newblock \emph{Future Generation Computer Systems}, 101:\penalty0 542--554, 2019.
\newblock ISSN 0167-739X.
\newblock \doi{https://doi.org/10.1016/j.future.2019.06.027}.
\newblock URL \url{https://www.sciencedirect.com/science/article/pii/S0167739X19306077}.

\bibitem[Roy et~al.(2019)Roy, Kiral-Kornek, and Harrer]{10.1007/978-3-030-21642-9_8}
Subhrajit Roy, Isabell Kiral-Kornek, and Stefan Harrer.
\newblock Chrononet: A deep recurrent neural network for abnormal eeg identification.
\newblock In \emph{Artificial Intelligence in Medicine: 17th Conference on Artificial Intelligence in Medicine, AIME 2019, Poznan, Poland, June 26–29, 2019, Proceedings}, page 47–56, Berlin, Heidelberg, 2019. Springer-Verlag.
\newblock ISBN 978-3-030-21641-2.
\newblock \doi{10.1007/978-3-030-21642-9_8}.
\newblock URL \url{https://doi.org/10.1007/978-3-030-21642-9_8}.

\bibitem[Biesmans et~al.(2017)Biesmans, Das, Francart, and Bertrand]{7478117}
Wouter Biesmans, Neetha Das, Tom Francart, and Alexander Bertrand.
\newblock Auditory-inspired speech envelope extraction methods for improved eeg-based auditory attention detection in a cocktail party scenario.
\newblock \emph{IEEE Transactions on Neural Systems and Rehabilitation Engineering}, 25\penalty0 (5):\penalty0 402--412, 2017.
\newblock \doi{10.1109/TNSRE.2016.2571900}.

\bibitem[Aboalayon et~al.(2016)Aboalayon, Faezipour, Almuhammadi, and Moslehpour]{e18090272}
Khald Ali~I. Aboalayon, Miad Faezipour, Wafaa~S. Almuhammadi, and Saeid Moslehpour.
\newblock Sleep stage classification using eeg signal analysis: A comprehensive survey and new investigation.
\newblock \emph{Entropy}, 18\penalty0 (9), 2016.
\newblock ISSN 1099-4300.
\newblock \doi{10.3390/e18090272}.
\newblock URL \url{https://www.mdpi.com/1099-4300/18/9/272}.

\bibitem[Liu et~al.(2023)Liu, Zhang, Hou, Mian, Wang, Zhang, and Tang]{9462394}
Xiao Liu, Fanjin Zhang, Zhenyu Hou, Li~Mian, Zhaoyu Wang, Jing Zhang, and Jie Tang.
\newblock Self-supervised learning: Generative or contrastive.
\newblock \emph{IEEE Transactions on Knowledge and Data Engineering}, 35\penalty0 (1):\penalty0 857--876, 2023.
\newblock \doi{10.1109/TKDE.2021.3090866}.

\bibitem[Wang et~al.(2024{\natexlab{a}})Wang, Zhao, Luo, Zhou, Jiang, Li, Li, and Pan]{wang2024cbramod}
Jiquan Wang, Sha Zhao, Zhiling Luo, Yangxuan Zhou, Haiteng Jiang, Shijian Li, Tao Li, and Gang Pan.
\newblock Cbramod: A criss-cross brain foundation model for eeg decoding.
\newblock \emph{arXiv preprint arXiv:2412.07236}, 2024{\natexlab{a}}.

\bibitem[Jiang et~al.(2024)Jiang, Zhao, and Lu]{ICLR2024_47393e85}
Wei-Bang Jiang, Liming Zhao, and Bao-liang Lu.
\newblock Large brain model for learning generic representations with tremendous eeg data in bci.
\newblock In B.~Kim, Y.~Yue, S.~Chaudhuri, K.~Fragkiadaki, M.~Khan, and Y.~Sun, editors, \emph{International Conference on Learning Representations}, volume 2024, pages 16405--16426, 2024.

\bibitem[Cecotti et~al.(2025)Cecotti, Shah, Jagadish, and Tanaka]{cecotti2025non}
Hubert Cecotti, Rashmi~Mrugank Shah, Raksha Jagadish, and Toshihisa Tanaka.
\newblock Non-stationarity in brain-computer interfaces: An analytical perspective.
\newblock \emph{arXiv preprint arXiv:2512.15941}, 2025.

\bibitem[Shen and Lin(2019)]{shen2019challenge}
Yi-Wei Shen and Yuan-Pin Lin.
\newblock Challenge for affective brain-computer interfaces: non-stationary spatio-spectral eeg oscillations of emotional responses.
\newblock \emph{Frontiers in human neuroscience}, 13:\penalty0 366, 2019.

\bibitem[Raza et~al.(2019)Raza, Rathee, Zhou, Cecotti, and Prasad]{RAZA2019154}
Haider Raza, Dheeraj Rathee, Shang-Ming Zhou, Hubert Cecotti, and Girijesh Prasad.
\newblock Covariate shift estimation based adaptive ensemble learning for handling non-stationarity in motor imagery related eeg-based brain-computer interface.
\newblock \emph{Neurocomputing}, 343:\penalty0 154--166, 2019.
\newblock ISSN 0925-2312.
\newblock \doi{https://doi.org/10.1016/j.neucom.2018.04.087}.
\newblock URL \url{https://www.sciencedirect.com/science/article/pii/S0925231219301560}.

\bibitem[Wang et~al.(2024{\natexlab{b}})Wang, Liu, He, Xu, Ma, and Li]{NEURIPS2024_4540d267}
Guagnyu Wang, Wenchao Liu, Yuhong He, Cong Xu, Lin Ma, and Haifeng Li.
\newblock Eegpt: Pretrained transformer for universal and reliable representation of eeg signals.
\newblock In A.~Globerson, L.~Mackey, D.~Belgrave, A.~Fan, U.~Paquet, J.~Tomczak, and C.~Zhang, editors, \emph{Advances in Neural Information Processing Systems}, volume~37, pages 39249--39280. Curran Associates, Inc., 2024{\natexlab{b}}.
\newblock \doi{10.52202/079017-1239}.
\newblock URL \url{https://proceedings.neurips.cc/paper_files/paper/2024/file/4540d267eeec4e5dbd9dae9448f0b739-Paper-Conference.pdf}.

\bibitem[Yang et~al.(2023)Yang, Westover, and Sun]{NEURIPS2023_f6b30f3e}
Chaoqi Yang, M~Westover, and Jimeng Sun.
\newblock Biot: Biosignal transformer for cross-data learning in the wild.
\newblock In A.~Oh, T.~Naumann, A.~Globerson, K.~Saenko, M.~Hardt, and S.~Levine, editors, \emph{Advances in Neural Information Processing Systems}, volume~36, pages 78240--78260. Curran Associates, Inc., 2023.
\newblock URL \url{https://proceedings.neurips.cc/paper_files/paper/2023/file/f6b30f3e2dd9cb53bbf2024402d02295-Paper-Conference.pdf}.

\bibitem[Hoi et~al.(2021)Hoi, Sahoo, Lu, and Zhao]{HOI2021249}
Steven~C.H. Hoi, Doyen Sahoo, Jing Lu, and Peilin Zhao.
\newblock Online learning: A comprehensive survey.
\newblock \emph{Neurocomputing}, 459:\penalty0 249--289, 2021.
\newblock ISSN 0925-2312.
\newblock \doi{https://doi.org/10.1016/j.neucom.2021.04.112}.
\newblock URL \url{https://www.sciencedirect.com/science/article/pii/S0925231221006706}.

\bibitem[Chen et~al.(2022)Chen, Wang, Darrell, and Ebrahimi]{9880363}
Dian Chen, Dequan Wang, Trevor Darrell, and Sayna Ebrahimi.
\newblock Contrastive test-time adaptation.
\newblock In \emph{2022 IEEE/CVF Conference on Computer Vision and Pattern Recognition (CVPR)}, pages 295--305, 2022.
\newblock \doi{10.1109/CVPR52688.2022.00039}.

\bibitem[Huang et~al.(2025)Huang, Li, Li, Wang, Zeng, Liang, Wu, Chen, Li, and Wang]{huang2025online}
Zhenpeng Huang, Xinhao Li, Jiaqi Li, Jing Wang, Xiangyu Zeng, Cheng Liang, Tao Wu, Xi~Chen, Liang Li, and Limin Wang.
\newblock Online video understanding: Ovbench and videochat-online.
\newblock In \emph{Proceedings of the Computer Vision and Pattern Recognition Conference}, pages 3328--3338, 2025.

\bibitem[Shi et~al.(2025)Shi, Xu, Wang, Qin, Wang, Wang, Wang, Ebrahimi, and Wang]{10.1145/3735633}
Haizhou Shi, Zihao Xu, Hengyi Wang, Weiyi Qin, Wenyuan Wang, Yibin Wang, Zifeng Wang, Sayna Ebrahimi, and Hao Wang.
\newblock Continual learning of large language models: A comprehensive survey.
\newblock \emph{ACM Comput. Surv.}, 58\penalty0 (5), November 2025.
\newblock ISSN 0360-0300.
\newblock \doi{10.1145/3735633}.
\newblock URL \url{https://doi.org/10.1145/3735633}.

\bibitem[Wang et~al.(2020)Wang, Shelhamer, Liu, Olshausen, and Darrell]{wang2020tent}
Dequan Wang, Evan Shelhamer, Shaoteng Liu, Bruno Olshausen, and Trevor Darrell.
\newblock Tent: Fully test-time adaptation by entropy minimization.
\newblock \emph{arXiv preprint arXiv:2006.10726}, 2020.

\bibitem[Jeon et~al.(2023)Jeon, Ko, Yoon, and Suk]{9508768}
Eunjin Jeon, Wonjun Ko, Jee~Seok Yoon, and Heung-Il Suk.
\newblock Mutual information-driven subject-invariant and class-relevant deep representation learning in bci.
\newblock \emph{IEEE Transactions on Neural Networks and Learning Systems}, 34\penalty0 (2):\penalty0 739--749, 2023.
\newblock \doi{10.1109/TNNLS.2021.3100583}.

\bibitem[Du et~al.(2022)Du, Xu, Wang, Liu, and Ma]{du2022eeg}
Yang Du, Yongling Xu, Xiaoan Wang, Li~Liu, and Pengcheng Ma.
\newblock Eeg temporal--spatial transformer for person identification.
\newblock \emph{Scientific Reports}, 12\penalty0 (1):\penalty0 14378, 2022.

\bibitem[Brunner et~al.(2008)Brunner, Leeb, M{\"u}ller-Putz, Schl{\"o}gl, and Pfurtscheller]{brunner2008bci}
Clemens Brunner, Robert Leeb, Gernot M{\"u}ller-Putz, Alois Schl{\"o}gl, and Gert Pfurtscheller.
\newblock Bci competition 2008--graz data set a.
\newblock \emph{Institute for knowledge discovery (laboratory of brain-computer interfaces), Graz University of Technology}, 16\penalty0 (1-6):\penalty0 1, 2008.

\bibitem[Steyrl et~al.(2016)Steyrl, Scherer, Faller, and M{\"u}ller-Putz]{steyrl2016random}
David Steyrl, Reinhold Scherer, Josef Faller, and Gernot~R M{\"u}ller-Putz.
\newblock Random forests in non-invasive sensorimotor rhythm brain-computer interfaces: a practical and convenient non-linear classifier.
\newblock \emph{Biomedical Engineering/Biomedizinische Technik}, 61\penalty0 (1):\penalty0 77--86, 2016.

\bibitem[Ma et~al.(2022)Ma, Yang, Qiu, Li, Gao, and Xia]{ma2022large}
Jun Ma, Banghua Yang, Wenzheng Qiu, Yunzhe Li, Shouwei Gao, and Xinxing Xia.
\newblock A large eeg dataset for studying cross-session variability in motor imagery brain-computer interface.
\newblock \emph{Scientific Data}, 9\penalty0 (1):\penalty0 531, 2022.

\bibitem[Goldberger et~al.(2000)Goldberger, Amaral, Glass, Hausdorff, Ivanov, Mark, Mietus, Moody, Peng, and Stanley]{goldberger2000physiobank}
Ary~L Goldberger, Luis~AN Amaral, Leon Glass, Jeffrey~M Hausdorff, Plamen~Ch Ivanov, Roger~G Mark, Joseph~E Mietus, George~B Moody, Chung-Kang Peng, and H~Eugene Stanley.
\newblock Physiobank, physiotoolkit, and physionet: components of a new research resource for complex physiologic signals.
\newblock \emph{circulation}, 101\penalty0 (23):\penalty0 e215--e220, 2000.

\bibitem[Schalk et~al.(2004)Schalk, McFarland, Hinterberger, Birbaumer, and Wolpaw]{schalk2004bci2000}
Gerwin Schalk, Dennis~J McFarland, Thilo Hinterberger, Niels Birbaumer, and Jonathan~R Wolpaw.
\newblock Bci2000: a general-purpose brain-computer interface (bci) system.
\newblock \emph{IEEE Transactions on biomedical engineering}, 51\penalty0 (6):\penalty0 1034--1043, 2004.

\bibitem[Chen et~al.(2023)Chen, Wang, Huang, Hu, Shen, and Zhang]{chen2023large}
Jingjing Chen, Xiaobin Wang, Chen Huang, Xin Hu, Xinke Shen, and Dan Zhang.
\newblock A large finer-grained affective computing eeg dataset.
\newblock \emph{Scientific Data}, 10\penalty0 (1):\penalty0 740, 2023.

\bibitem[Liu et~al.(2021)Liu, Qiu, Zheng, and Lu]{liu2021comparing}
Wei Liu, Jie-Lin Qiu, Wei-Long Zheng, and Bao-Liang Lu.
\newblock Comparing recognition performance and robustness of multimodal deep learning models for multimodal emotion recognition.
\newblock \emph{IEEE Transactions on Cognitive and Developmental Systems}, 14\penalty0 (2):\penalty0 715--729, 2021.

\bibitem[Hu et~al.(2022)Hu, Shen, Wallis, Allen-Zhu, Li, Wang, Wang, Chen, et~al.]{hu2022lora}
Edward~J Hu, Yelong Shen, Phillip Wallis, Zeyuan Allen-Zhu, Yuanzhi Li, Shean Wang, Liang Wang, Weizhu Chen, et~al.
\newblock Lora: Low-rank adaptation of large language models.
\newblock \emph{Iclr}, 1\penalty0 (2):\penalty0 3, 2022.

\end{thebibliography}
